\documentclass[10pt,conference]{IEEEtran}
\IEEEoverridecommandlockouts
\usepackage{cite}
\usepackage{amsmath,amssymb,amsfonts}
\usepackage{algorithmic}
\usepackage{graphicx}
\usepackage{textcomp}
\def\BibTeX{{\rm B\kern-.05em{\sc i\kern-.025em b}\kern-.08em
    T\kern-.1667em\lower.7ex\hbox{E}\kern-.125emX}}

\usepackage[utf8]{inputenc} % allow utf-8 input
\usepackage[T1]{fontenc}    % use 8-bit T1 fonts
\usepackage{url}            % simple URL typesetting
\usepackage{booktabs}       % professional-quality tables
\usepackage{nicefrac}       % compact symbols for 1/2, etc.
\usepackage{microtype}      % microtypography
\usepackage{listings}
\usepackage{comment}
\usepackage{multirow}
\usepackage{adjustbox}
\usepackage[table]{xcolor}
\usepackage{threeparttable}
\usepackage[caption=false,font=footnotesize]{subfig}
\usepackage{enumitem}
\usepackage[most]{tcolorbox}
\usepackage{lipsum} % for placeholder text
\usepackage{balance}
\tcbuselibrary{breakable}
\definecolor{highlightcolor}{gray}{0.90}
\lstdefinestyle{sparql}{
  language=SQL, % 用 SQL 接近 SPARQL 的结构
  basicstyle=\ttfamily\small,
  keywordstyle=\color{blue},
  stringstyle=\color{red},
  commentstyle=\color{gray},
  morekeywords={SELECT, WHERE}, % 手动添加 SPARQL 关键字
  frame=single,
  breaklines=true,
  columns=fullflexible
}
\IEEEaftertitletext{\vspace{-0.2cm}}

\begin{document}
\title{KG-Hopper: Empowering Compact Open LLMs with Knowledge Graph Reasoning via Reinforcement Learning}

\author{
\IEEEauthorblockN{Shuai Wang, \quad \quad Yinan Yu}
\IEEEauthorblockA{
  Department of Computer Science and Engineering \\
  Chalmers University of Technology and University of Gothenburg \\
  SE-41296 Gothenburg, Sweden \\
  \texttt{\{shuaiwa, yinan\}@chalmers.se}
}
}

\maketitle

\begin{abstract}
Large Language Models (LLMs) demonstrate impressive natural language capabilities but often struggle with knowledge-intensive reasoning tasks. Knowledge Base Question Answering (KBQA), which leverages structured Knowledge Graphs (KGs) exemplifies this challenge due to the need for accurate multi-hop reasoning. Existing approaches typically perform sequential reasoning steps guided by predefined pipelines, restricting flexibility and causing error cascades due to isolated reasoning at each step. To address these limitations, we propose KG-Hopper, a novel Reinforcement Learning (RL) framework that empowers compact open LLMs with the ability to perform integrated multi-hop KG reasoning  within a single inference round. Rather than reasoning step-by-step, we train a \textit{Reasoning LLM} that embeds the entire KG traversal and decision process into a unified “thinking” stage, enabling global reasoning over cross-step dependencies and dynamic path exploration with backtracking. Experimental results on eight KG reasoning benchmarks show that KG-Hopper, based on a 7B-parameter LLM, consistently outperforms larger multi-step systems (up to 70B) and achieves competitive performance with proprietary models such as GPT-3.5-Turbo and GPT-4o-mini, while remaining compact, open, and data-efficient. 
The code is publicly available at:
\url{https://github.com/Wangshuaiia/KG-Hopper}.
\end{abstract}


\section{Introduction}
Large Language Models (LLMs) have achieved remarkable success across diverse domains but still exhibit notable shortcomings, such as hallucinations and factual inaccuracies, particularly in knowledge-intensive tasks. This is largely due to the implicit storage of knowledge within their model weights, making knowledge updates cumbersome and resource-intensive through fine-tuning. To address this limitation, Retrieval-Augmented Generation (RAG) has emerged as an effective strategy, enabling LLMs to dynamically access external knowledge during inference \cite{wang2025plugging}. Among various external resources, Knowledge Graphs (KGs) stand out as structured and reliable knowledge bases, offering explicit, interpretable, and easily updateable knowledge beneficial in critical applications such as medicine and finance.

Knowledge Base Question Answering (KBQA), which aims to leverage structured knowledge from KGs to answer questions, often requires complex multi-hop reasoning—traversing multiple interconnected relationships within a KG \cite{wang-yu-2025-iquest}. Current approaches typically follow step-by-step reasoning strategies, sequentially processing entities and relationships from a predefined pipeline. For example, as illustrated in Figure \ref{fig:main}(a), answering the question "What is the official flower of the area affected by Tropical Storm Fabio?" involves first identifying the affected area and subsequently retrieving its official flower. However, such rigid frameworks present critical drawbacks:

\begin{figure*}[ht]
    \centering
    \includegraphics[width=\textwidth]{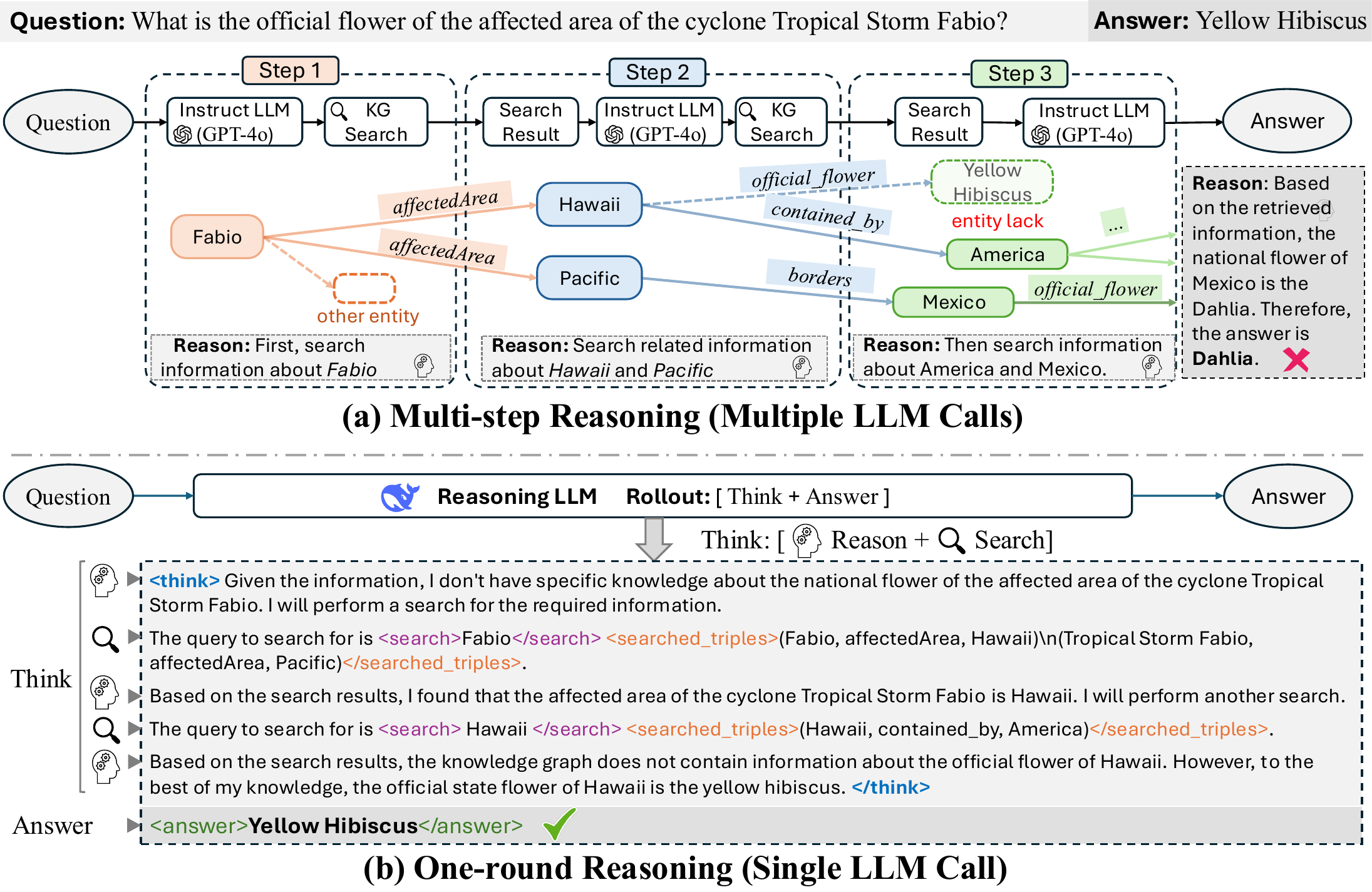} 
    \caption{Multi-step vs one-round multi-hop reasoning over a knowledge graph: \textbf{(a)} multi-step reasoning and \textbf{(b)} our one-round reasoning. The multi-step pipeline invokes multiple sequential LLM calls and fails due to the missing entity \textit{Yellow Hibiscus}, leading to an incorrect path. In contrast, our one-round approach performs the entire reasoning process within a \textit{single Reasoning LLM} call, maintaining coherence and demonstrating robustness to incomplete knowledge.}
    \label{fig:main}
\end{figure*}
\begin{itemize}[leftmargin=1.5em]
\item \textbf{Limited Flexibility and Local Optima Susceptibility.} Traditional methods sequentially aggregate information through individual inference steps guided by predefined reasoning paths. Consequently, these methods struggle to dynamically adjust when facing incomplete or misleading KG data. For instance, as shown in Figure \ref{fig:main}(a), missing the critical entity "Yellow Hibiscus" could mislead a multi-step method employing beam search into incorrectly choosing the national flower of Mexico, with limited capability for backtracking.
\item \textbf{Error Cascading and Reasoning Bias.} Stepwise methods inherently propagate errors from earlier steps. Incorrectly choosing a node, such as "Mexico" in Figure \ref{fig:main}(a), directly influences subsequent reasoning steps. Additionally, treating each reasoning step independently neglects inter-step dependencies, causing biases and potential deviations from the original query intent.
\end{itemize}
To mitigate these limitations, two lines of research offer promising advances. First, Reinforcement learning (RL) have demonstrated its effectiveness in navigating discrete and combinatorial decision spaces, such as KG traversal by optimizing reasoning policies through exploration and long-term rewards~\cite{xiong2017deeppath, wang2026domagent, lin2018multi}. % By optimizing reasoning policies through trial-and-error interactions, RL enables models to explore multiple candidate paths, making it particularly suitable for multi-hop KG reasoning tasks.
Second, recent work on \textit{Reasoning LLMs}, such as ChatGPT-o1~\cite{jaech2024openai} and DeepSeek-R1~\cite{guo2025deepseek}, has significantly improved the reasoning capabilities of language models. These models perform a dedicated “thinking” phase before generating answers, where each token in the thought sequence can be viewed as a latent variable. Importantly, later tokens can revise earlier ones, enabling self-reflection and correction within the reasoning process \cite{zhu2025chain}. This iterative refinement mechanism aligns naturally with the multi-hop nature of complex KBQA tasks, allowing LLMs to reason more coherently and correct potential errors before answer generation.

Building on these insights, we propose \textbf{KG-Hopper}, a novel KBQA framework integrating the entire multi-hop reasoning process within a single LLM call, as depicted in Figure \ref{fig:main}(b). Specifically, KG-Hopper leverages RL to enhance the reasoning capabilities of LLMs by training them to utilize KG retrieval tools effectively.

KG-Hopper addresses previous limitations with the following advantages:
\textbf{(1)} By embedding the complete reasoning chain within a single inference step, KG-Hopper captures cross-step dependencies, mitigating local biases and ensuring coherent reasoning.
\textbf{(2)} Integrating reasoning into the \textit{Reasoning LLM}'s "thinking" phase allows flexible exploration and effective backtracking, significantly reducing cascading errors without predefined reasoning patterns.
\textbf{(3)} Our framework leverages RL to explicitly guide and refine reasoning processes, substantially improving the model’s performance in complex multi-hop scenarios.
\textbf{(4)} It remains compact and deployable, using a 7B LLM to achieve competitive results with much larger models.

To effectively train KG-Hopper, we initially generate cold-start data and employ Supervised Fine-Tuning (SFT), enabling autonomous KG retrieval. We subsequently strengthen multi-hop reasoning capabilities through targeted RL training, guided by carefully designed reward functions for retrieving and reasoning process. Additionally, we mask the retrieved triples from KG to minimize external knowledge interference and employ history resampling to enhance training efficiency.

Our main contributions are as follows:
\begin{itemize}[leftmargin=1.5em]
% \item We integrate multi-hop reasoning over KGs directly within the "thinking" phase of a reasoning LLM, enabling the model to autonomously and iteratively retrieve necessary knowledge from the KG within a single inference round.

\item We introduce a RL framework for LLM–KG integration, combining structured reward signals for retrieval and reasoning with masked supervision over KG content. To the best of our knowledge, this is the first work to apply RL to enable end-to-end KG reasoning within LLMs.

\item We propose a one-round KBQA framework that performs multi-hop reasoning entirely within the “thinking” phase of a single LLM inference, enabling autonomous and iterative KG traversal without multi-step orchestration.

\item We achieve strong empirical results on eight KBQA benchmarks: our compact 7B-parameter model consistently outperforms multi-step methods using models up to 70B, and matches or exceeds the performance of proprietary models such as GPT-3.5-Turbo and GPT-4o-mini.
\end{itemize}

\section{Task Definition}

We address the task of Knowledge Base Question Answering (KBQA) over a knowledge graph $G = \{(e, r, e') \mid e, e' \in \mathcal{E}, r \in \mathcal{R} \}$, where each triple $(e, r, e')$ denotes a relation $r$ between two entities $e$ and $e'$. Given a natural language question $q$, along with an identified topic entity $e_t \in \mathcal{E}$ mentioned in the question, the objective is to infer the correct answer entity $e_a \in \mathcal{E}$ by reasoning over the knowledge graph $G$.
The answering process typically starts from the topic entity $e_t$ and involves exploring the graph to identify the answer entity $e_a$. In practice, large-scale knowledge graphs are often sparse and incomplete, making it difficult to locate the answer through simple one-hop queries. As a result, multi-hop reasoning is frequently required, where the answer entity $e_a$ may reside several hops away from the topic entity $e_t$.

\section{Method}
Our approach leverages reinforcement learning to train LLMs for KG retrieval and reasoning. This section introduces two key components: the construction of a knowledge graph retrieval tool, and the RL framework for training LLMs to interact with the knowledge graph.

\subsection{Knowledge Graph Retrieval Tool}

Large-scale knowledge graphs are typically stored in graph databases and queried using SPARQL for retrieval. Our retrieval process starts from a topic entity and expands the search. Due to the vast number of triples connected to a single node, it is crucial to filter out irrelevant information. To address this, we adopt a two-stage strategy: first retrieving the set of directly connected edges (predicates) and then selecting the ones relevant to the input query.

Given a topic entity, our tool first retrieves all predicates and corresponding object entities linked to it. For example, given the question \textit{``What books did J.K. Rowling write?''}, the topic entity is \texttt{J.K. Rowling}. The following SPARQL query is issued to retrieve all outgoing predicates and their objects:

\begin{lstlisting}[style=sparql]
SELECT ?predicate ?object
WHERE {
  ns:m.05b6w ?predicate ?object .
}
\end{lstlisting}

\noindent
Here, \verb|ns:m.05b6w| denotes the Freebase ID for J.K. Rowling, \verb|?predicate| retrieves all relations where she appears as the subject, and \verb|?object| returns the corresponding object entities. 

For example, the retrieved predicates include: \\
\hspace*{1.5em} \verb|ns:book.author.works_written|, \\
\hspace*{1.5em} \verb|ns:people.person.place_of_birth|, \\
\hspace*{1.5em} \verb|ns:people.person.nationality|. 

Based on the semantic relevance to the question, the most related predicate, such as \verb|ns:book.author.works_written|, is selected. This predicate indicates the books authored by the subject.
Next, the tool issues another SPARQL query to retrieve the tail entities (i.e., books) associated with the selected predicate:

\begin{lstlisting}[style=sparql]
SELECT ?tailEntity
WHERE {
  ns:m.05b6w ns:book.author.works_written ?tailEntity .
}
\end{lstlisting}

\noindent
This query fetches all book entities linked to J.K. Rowling through the \texttt{works\_written} relation, effectively answering the question.

In summary, the retrieval tool takes an entity as input and returns a set of relevant triples from the knowledge graph by identifying and filtering meaningful relations and associated entities.
% 而后，由于获取足够的long CoT reasoning data非常困难, 因此使用RLHF来激发和提升模型的解决多步推理问题的能力。

\subsection{Cold Start}

% Before introducing reinforcement learning to enhance the model's reasoning capability, we follow the strategy inspired by DeepSeek-R1 \cite{guo2025deepseek} 
To avoid the unstable cold-start phase typically seen in early RL training. Specifically, we construct and collect a small set of CoT annotated data to fine-tune the base LLM, which is then used as the initial RL actor. The cold-start dataset is created with two primary objectives: (1) to demonstrate how the model should properly invoke the knowledge graph retrieval tool; and (2) to enforce a consistent, structured format for answer generation.

We define a rule such that when the LLM generates the special tokens \texttt{<search>} and \texttt{</search>}, it triggers the KG retrieval tool to search the enclosed entity. The retrieved triples are then wrapped with \texttt{<searched\_triples>} tags and appended to the current context, enabling the LLM to continue generation with access to the retrieved knowledge.
To collect cold start data, we use few-shot prompting with a long CoT example to elicit responses from a powerful LLM. We then select examples that correctly invoke KG queries and exhibit high readability. 
For instance, given the question:
\textit{what timezone is Utah in?}
A preferred response would begin with an explicit motivation for search, such as:

\textit{\textcolor{blue}{<think>}Given the information, I don't have specific knowledge about Utah's timezone. I will perform a search for the required information. The query to search for is \textcolor{purple}{<search>}utah\textcolor{purple}{</search>} \textcolor{orange}{<searched\_triples>} (Utah, timeZone, Mountain Time Zone)\textcolor{orange}{</searched\_triples>}. Based on the search results, I found that Utah is in the Mountain Time Zone.\textcolor{blue}{</think>} \textcolor{teal}{<answer>}Mountain Standard Time\textcolor{teal}{</answer>}}

We select such examples, which exhibit clarity, appropriate tool invocation, and well-structured reasoning, as the cold-start training data. Such structure is the desired output format of the LLM. The \texttt{<think>} section encapsulates the full CoT reasoning process, including autonomous invocation of the KG tool (\texttt{<search>}), retrieval results (\texttt{<triples>}), and intermediate deductions. The \texttt{<answer>} section then summarizes the reasoning into a final answer. This also represents the desired output format of the LLM. 

We use the collected data to fine-tune the base LLM. During training, we mask the tokens within the \texttt{<triples>} to prevent the model from being distracted by retrieved knowledge. This encourages the model to generalize its reasoning strategy while preserving tool-use behaviors. The fine-tuned model is then further optimized via reinforcement learning.

\subsection{Reasoning-oriented Reinforcement Learning}
% 这里写上prompt
Given the difficulty of obtaining sufficient high-quality long CoT reasoning data, we leverage Reinforcement Learning with Human Feedback (RLHF) as a principled alternative to explicitly guide the model toward effective multi-hop reasoning. 
We design a composite reward function to guide the model's behavior throughout the reasoning process. The reward comprises four components: retrieval reward, format reward, reasoning reward, and final answer reward.
\begin{itemize}[leftmargin=1.5em]
\item \textbf{Retrieval Reward}
To encourage the model to search for answers through the knowledge graph rather than relying solely on its internal knowledge, we provide a positive reward for each invocation of the query tool. However, to prevent the model from overusing the tool solely for reward accumulation, we apply a cap on the maximum reward. The retrieval reward is defined as:
\begin{equation}
R_{\text{search}} = \min(0.5 \cdot n, 0.8)
\end{equation}
where $n$ denotes the number of times the query tool is invoked. This design incentivizes query usage while discouraging excessive or redundant retrievals.

\item \textbf{Format Reward}
% 这里我们的目的是促进模型在回答问题之前可以进行推理，并且在推理的过程中可以调用方程
To enforce structured reasoning and tool usage, we define a format reward that requires the generated text to follow a predefined format. Specifically, the model must use the tags \texttt{<think>}, \texttt{<search>}, and \texttt{<answer>} appropriately. If all tags appear in the correct positions and order, a fixed reward is granted:
\begin{equation}
R_{\text{format}} = 
\begin{cases}
0.5 & \text{if the format meets all requirements} \\
0   & \text{otherwise}
\end{cases}
\end{equation}
This constraint ensures that the model explicitly separates its reasoning, search actions, and final answer.

\item \textbf{Reasoning Reward}
In multi-hop KG reasoning, an error at any intermediate step can lead to an incorrect final answer.  To mitigate this, we introduce a reward signal that directly evaluates the quality of the reasoning process itself, encouraging the model to make sound decisions at each step, rather than focusing solely on the final answer.
% To address this, we aim to guide the reasoning process such that the model maximizes its reward for each intermediate step. 
% If the retrieved triples provide sufficient information, the model is expected to organize and synthesize an answer. If not, it should perform further reasoning by selecting the next entity or relation based on the current knowledge. In cases where the knowledge graph lacks the necessary information (due to incompleteness), the model may fall back on its own background knowledge.
The model is expected to adapt its behavior based on the informativeness of retrieved triples. When the retrieved information is sufficient, the model should organize and synthesize it into a coherent answer. Otherwise, it should continue reasoning by selecting the next entity or relation to query. In cases where the KG lacks the necessary facts (e.g., due to incompleteness), the model is allowed to fall back on its internal knowledge to infer a plausible answer.
% Added text but too long: To assess whether the model makes a reasonable decision based on the retrieved knowledge, we employ a separate LLM to evaluate the reasoning process encapsulated within the \texttt{<think>} tag. 
To assess the quality of the model’s reasoning behavior, we use an external LLM to evaluate the full reasoning trace enclosed in the \texttt{<think>} tag. % This evaluation provides a scalar reward indicating how logically consistent, well-grounded, and goal-directed the reasoning sequence is. The resulting score constitutes the reasoning reward and serves as an essential signal for guiding long-horizon inference.
The reasoning reward is computed as:
\begin{equation}
R_{\text{reason}} = f_r(\text{reasoning process}) \in (0, 1)
\label{eq:reason}
\end{equation}
where $f_r$ denotes an external LLM, and a higher score indicates a more reasonable and logically valid reasoning trace.

\item \textbf{Answer Reward}
Finally, we reward the model if it provides a correct answer. Since the generated answer may differ from the ground truth due to variations such as abbreviations or phrasing, we again use a separate LLM to perform semantic similarity assessment between the predicted answer and the ground truth (\textbf{LLM as Evaluator}). The final answer reward is defined as:
\begin{equation}
R_{\text{answer}} = f_a(\text{predicted\_answer}, \text{ground\_truth}) \in \{0, 1\}
\label{eq:answer}
\end{equation}
where $f_a$ denotes an external LLM, and the predicted answer is extracted from the \texttt{<answer>} tag. A reward of 1 is given if the LLM determines the answer is semantically correct, and 0 otherwise.
In the absence of ground truth, we directly use a (preferably larger) LLM with rich knowledge as both the \textbf{Judge} and \textbf{Evaluator}, i.e., we rely on the LLM’s internal knowledge to assess whether the predicted answer is correct.
\end{itemize}
By combining the above four reward functions, the total reward provides comprehensive guidance to the model throughout the reasoning process, promoting structured, accurate, and knowledge-grounded answers, which can be represented as:
\begin{equation}
R_{\text{final}} = R_{\text{search}} + R_{\text{format}} + R_{\text{reason}} + R_{\text{answer}}  
\end{equation}

\subsection{Optimization}

We optimize the reasoning policy using Group Relative Policy Optimization (GRPO), which trains the LLM to maximize the expected reward of generated reasoning trajectories. Formally, the objective is simplified as:
\begin{equation}
\theta^{*} = \arg\max_{\theta} \; \mathbb{E}_{q \sim \pi_{\theta}} 
\big[ R_{\text{final}}(q) \big],
\end{equation}
where $\pi_{\theta}$ denotes the LLM policy and $R_{\text{final}}(q)$ is the corresponding reward function. GRPO estimates relative advantages within sampled output groups to stabilize optimization and encourage higher-quality reasoning paths.

\textbf{Masking Retrieved Triples.}
Retrieved triples are provided as auxiliary context but are not expected to be generated by the model. We therefore mask tokens enclosed by \texttt{<triples>} and \texttt{</triples>} during loss computation to prevent the model from learning to reproduce retrieved content.

\textbf{History Resampling for Efficient Training.}
In KBQA, simple one-hop queries often produce uniformly high rewards, leading to near-zero normalized advantages and inefficient learning. Following a history resampling strategy~\cite{zhang2025srpo}, we remove one-hop questions after an initial training phase, encouraging the model to focus on multi-hop reasoning in a curriculum learning manner~\cite{narvekar2020curriculum}.

\section{Experiments}
\subsection{Datasets}
\label{sec:datasets-experimens}
We evaluate our approach on eight widely-used datasets for KBQA, leveraging two large-scale general-purpose knowledge graphs: Freebase and WikiData. Four of the datasets are based on the Freebase knowledge graph and are standard KBQA benchmarks: ComplexWebQuestions (CWQ)~\cite{talmor2018web}, WebQuestionsSP (WebQSP)~\cite{yih2016value}, WebQuestions~\cite{berant2013semantic}, and GrailQA~\cite{gu2021beyond}. The other four datasets are grounded in the WikiData knowledge graph. Among them, QALD10-en \cite{perevalov2022qald} is a KBQA dataset, while T-REx \cite{elsahar2018t} and Zero-Shot RE \cite{petroni2021kilt} are designed for slot filling tasks, and Creak \cite{onoe2creak} focuses on factual verification. 
Following prior work~\cite{sun2024thinkongraph, zhao2024kg, xiong-etal-2024-interactive}, we use Hit@1 score as the evaluation metric for both question answering and slot filling tasks, while accuracy is used to evaluate performance on the Creak dataset. 

\subsection{Implementation Details}
We conduct experiments using two instruction-tuned language models as backbones: LLaMA-3.1-8B-Instruct and Qwen-2.5-7B. To mitigate the unstable cold-start phase of reinforcement learning, we first construct a set of 500 high-quality examples to teach the models how to properly invoke the knowledge retrieval tool. We randomly sample 2,000 examples from 8 datasets for RL training. The scoring model $f_r$ for the reasoning process is Llama-3.3-70B, while the model $f_a$ used to determine whether the predicted answer matches the ground truth is Llama-3.2-3B.
Starting from the second epoch, we apply a resampling strategy to dynamically filter out trivial questions and retain more informative ones for continued training.
The training is performed on 8 NVIDIA A100 80G GPUs in total. For each input query, we generate 16 outputs (rollouts). We train for 2 epochs with a batch size of 16 and a learning rate of $1\text{e}{-6}$. The rollout temperature is set to 1, the PPO clip ratio is 0.2, and the KL divergence penalty coefficient is $1\text{e}{-5}$.

\begin{table*}[ht]
\renewcommand{\arraystretch}{1.2} 
\centering
\vspace{-0.5cm}
\caption{LLM+KG integration techniques. We highlight our focus in green.}
\label{tab:llm_kg_dimensions}
\resizebox{\textwidth}{!}{
\begin{tabular}{p{3cm}p{6cm}p{6cm}p{8cm}}
\hline
\textbf{Method} & \textbf{Short Description} & \textbf{Pros} & \textbf{Cons} \\
\hline
\rowcolor{gray!10} 
\multicolumn{4}{l}{\textbf{Model-KG Integration Strategies}}\\
Prompt-only LLM & LLM answers with parametric knowledge only & Easy to use; no dependencies & Poor on multi-hop or unseen queries \\
LLM + KG (no training) & Uses KG retrieval tools, no adaptation & Quick to integrate KG; flexible & No reasoning adaptation; tool orchestration needed \\
LLM + KG (SFT) & LLM fine-tuned with KG reasoning traces & More accurate; learns tool use & Requires labeled CoT data; brittle over long chains \\
\rowcolor{green!20} 
{LLM + KG (RL)} & {RL-trained LLM with KG traversal tools} & {Strong generalization; fewer examples needed} & {Requires reward design} \\
\hline
\rowcolor{gray!10}
\multicolumn{4}{l}{\textbf{Inference and Reasoning Modes}} \\
Direct prompting & Direct answer generation in one pass & Fastest inference; minimal cost & Cannot model intermediate reasoning \\
Multi-step reasoning & Sequential steps with intermediate tool calls & Fine-grained control; interpretable traces & High latency; error-prone; complex logic needed \\
\rowcolor{green!20} 
One-round reasoning & Full reasoning in one pass with embedded retrieval & Enables globally coherent reasoning; low latency & Requires strong reasoning capability; requires model adaptation \\
\hline
\rowcolor{gray!10}
\multicolumn{4}{l}{\textbf{Post-hoc Ranking and Aggregation}} \\
\rowcolor{green!20}
Node ranking & Rank multiple answer node candidates & Boosts precision when candidates are relevant & No explicit reasoning path; requires candidate set \\
Path ranking & Rank and select from reasoning chains & More robust to noise or ambiguity & High compute cost; poor scalability at runtime \\
\hline
\rowcolor{gray!10}
\multicolumn{4}{l}{\textbf{LLM Source and Accessibility}} \\
\rowcolor{green!20}
Open LLM & Models such as LLaMA, Qwen, DeepSeek & Transparent; reproducible; modifiable; cheaper to deploy & May underperform on some benchmarks without adaptation \\
Proprietary LLM & Closed-source APIs like GPT-4o & Often stronger zero-shot reasoning & Expensive; limited control; no model access; data privacy issues \\
\hline
\end{tabular}}
\end{table*}

\subsection{Main Results}
\label{sec:main-results}
We design our experiments to explore the reasoning effectiveness, training efficiency, and design trade-offs of our method KG-Hopper. 
To contextualize our approach within the broader KBQA landscape, Table~\ref{tab:llm_kg_dimensions} summarizes different LLM+KG integration strategies. It highlights variations across model integration methods, inference modes, post-processing techniques, and model accessibility. 
KG-Hopper is designed as an open-source, compact, and flexible solution that promotes transparency in the training process and supports targeted modifications. Through our experiments, we investigate how RL contributes to efficient multi-hop KBQA. Results are reported in Table~\ref{tab:grouped-llm-comparison}.

In terms of integration strategy, prompt-only LLMs, including strong proprietary models like GPT-4o, consistently underperform on multi-hop KBQA tasks. This highlights the limitations of relying solely on parametric knowledge and the lack of explicit, structured reasoning. Adding KG retrieval tools improves performance significantly (often by 10–30 Hits@1), but without any form of model-level adaptation, these systems exhibit fixed reasoning patterns and struggle to generalize beyond shallow queries. SFT on KG reasoning tasks further improves results, especially for moderately complex questions. However,  its imitation-based learning process tends to be brittle, it teaches the model to reproduce specific reasoning paths rather than adaptively exploring alternatives. Our RL-trained model, {KG-Hopper}, consistently outperforms SFT-based models of similar (7-8B) or larger size (13B, 70B) and matches or exceeds the performance of GPT-4o-mini + KG, particularly on more complex multi-hop reasoning tasks.

\begin{table*}[ht]
\centering
% \vspace{-1cm}
\caption{
% Comparison of different methods across multiple benchmarks. 
Performance comparison (Hits@1). Results marked with `*' are taken directly from the corresponding original papers. 
Bold numbers indicate the \textbf{best performance}, while underlined numbers denote the \underline{second-best}.
% For fair comparison, results from methods based on GPT-4 or GPT-4o are provided separately in Table~\ref{tab:results-gpt4} (Hits@1).
}
\centering
\renewcommand{\arraystretch}{1}
\begin{adjustbox}{width=\textwidth}
\begin{tabular}{lccccccccc}
\toprule
\multirow{2}{*}{\textbf{Method}} & \multirow{2}{*}{\textbf{Size}} & \multicolumn{4}{c}{\textbf{Freebase}} & \multicolumn{4}{c}{\textbf{WikiData}} \\
\cmidrule(lr){3-6} \cmidrule(lr){7-10}
 & & CWQ & WebQSP & WebQuestion & GrailQA & QALD10-en & T-REx & Zero-Shot RE & Creak \\
\midrule
\rowcolor{gray!10} \multicolumn{10}{l}{\textbf{LLM Prompting Only}} \\
Qwen-2.5-7B & 7B & 31.25 & 46.97 & 44.23 & 29.53 & 41.88 & 31.15 & 7.84 & 73.26 \\
LLaMA-3.1-8B & 8B & 32.33 & 45.07 & 45.88 & 28.35 & 40.25 & 23.00 & 12.54 & 75.80 \\
LLaMA-3.3-70B & 70B & 37.20 & 71.12 & 59.73 & 33.79 & 56.00 & 20.12 & 18.55 & 83.72 \\
DeepSeek-R1-Distill-Llama-70B & 70B & 31.92 & 77.43 & 68.84 & 31.70 & 43.10 & 34.21 & 22.27 & 79.10 \\
GPT-4o-mini & – & 42.32 & 65.97 & 57.26 & 36.22 & 51.98 & 26.90 & 18.85 & 83.72 \\
GPT-4o & – & 41.77 & 72.55 & 64.79 & 35.01 & 56.20 & 44.46 & 48.20 & 90.70 \\
\midrule
\rowcolor{gray!10}\multicolumn{10}{l}{\textbf{KG-Augmented LLMs without Fine-Tuning}} \\
Qwen-2.5-7B + KG & 7B & 44.82 & 72.60 & 56.33 & 41.10 & 56.54 & 64.21 & 70.32 & 79.04 \\
LLaMA-3.1-8B + KG & 8B & 45.64 & 71.32 & 57.05 & 40.40 & 55.73 & 65.80 & 68.66 & 80.50 \\
LLaMA-3.3-70B + KG & 70B & 44.00 & 81.90 & 72.60 & 57.70 & 71.78 & 65.42 & \textbf{80.42} & 87.04 \\
DeepSeek-R1-Distill-Llama-70B + KG & 70B & 52.38 & 81.34 & \underline{75.80} & \underline{59.02} & 66.10 & 72.04 & 78.04 & \underline{91.20} \\
GPT-4o-mini + KG & – & 54.35 & \textbf{84.40} & \textbf{81.02} & \textbf{60.00} & \underline{72.86} & 69.70 & 75.12 & 90.20 \\
KG-CoT w/GPT 3.5-Turbo~\cite{zhao2024kg} & – & 51.6$^{*}$ & 82.1$^{*}$ & 66.5$^{*}$ & - & - & - & - & - \\
% ToG w/GPT 3.5-Turbo~\cite{sun2024thinkongraph} & – & 57.1$^{*}$ & 76.2$^{*}$ & 68.7$^{*}$ & 50.2$^{*}$ & 54.5$^{*}$ & \textbf{76.8$^{*}$} & \textbf{88.0$^{*}$} & \underline{91.2$^{*}$} \\
\midrule
\rowcolor{gray!10}\multicolumn{10}{l}{\textbf{KG-Augmented LLMs with Supervised Fine-Tuning}} \\
Interactive-KBQA w/LLaMA-7B~\cite{xiong-etal-2024-interactive} & 7B & 39.9$^{*}$ & 43.57$^{*}$ & - & - & - & - & - & - \\
KG-CoT w/LLaMA-7B~\cite{zhao2024kg} & 7B & 46.7$^{*}$ & 72.4$^{*}$ & - & - & - & - & - & - \\
Qwen-2.5-7B (SFT) + KG & 7B & 51.84 & 74.80 & 61.42 & 46.18 & 65.18 & 68.77 & 71.46 & 83.43 \\
LLaMA-3.1-8B (SFT) + KG & 8B & 47.40 & 72.98 & 60.00 & 47.70 & 59.63 & 70.23 & 64.08 & 84.47 \\
Interactive-KBQA w/LLaMA-13B~\cite{xiong-etal-2024-interactive} & 13B & 42.5$^{*}$ & 54.86$^{*}$ & - & - & - & - & - & - \\
KG-CoT w/LLaMA-13B~\cite{zhao2024kg} & 13B & 50.0$^{*}$ & 74.6$^{*}$ & - & - & - & - & - & - \\
% \midrule
% \rowcolor{gray!10}\multicolumn{10}{l}{\textbf{KG-Augmented LLM w/ GPT-3.5-Turbo}} \\
\midrule
\rowcolor{gray!10}\multicolumn{10}{l}{\textbf{KG-Augmented LLMs with RL Fine-Tuning}} \\
{KG-Hopper w/Qwen-2.5-7B (ours)} & 7B & \textbf{61.07} & \underline{83.20} & 66.90 & 50.10 & \textbf{74.28} & \textbf{72.14} & \underline{78.64} & \textbf{91.82} \\
% \textbf{LLaMA-3.1-8B (RL) + KG} & 8B & \underline{58.20} & 76.90 & 67.28 & 55.41 & 67.66 & \underline{74.20} & 70.25 & 88.31 \\
% {KG-Hopper w/Qwen-2.5-7B (GPT4o guided $f_{r}$)} & 7B & 62.45 & 83.67 & 67.35 & 49.50 & 75.02 & 71.60 & 79.23 & 91.05 \\
\bottomrule
\end{tabular}%
\end{adjustbox}
\label{tab:grouped-llm-comparison}
\end{table*}

\subsection{Ablation}

\paragraph{RL vs SFT.} How does reinforcement learning compare to SFT in the context of multi-hop KBQA performance?
% and sample efficiency? NOTES: Maybe we can run some ablation here
(\textbf{RQ1})
Table~\ref{tab:sft_vs_rl} reports the Hits@1 scores across eight KBQA datasets, comparing models trained with SFT and RL using two LLM backbones: Qwen-2.5-7B and LLaMA-3.1-8B. In both cases, the RL-trained models consistently outperform their SFT counterparts, achieving gains of +4\% to +10\%. 
This performance gap can be attributed to the inherent misalignment issue \cite{wang2024math} in SFT, where the model learns to produce fixed, pattern-based responses. Such rigidity often leads the model to rely on spurious correlations from training examples, which can interfere with its ability to reason based on its own knowledge. In contrast, RL allows the model to adaptively coordinate with the retrieval module, perform more coherent multi-hop reasoning, and explicitly generate step-by-step solutions. 

Notably, the performance gains are especially significant on complex datasets such as CWQ and QALD10-en, with improvements of around 10\%. These datasets demand longer reasoning chains and deeper traversal of the KG. 
The results illustrate the strength of RL in capturing long-horizon dependencies, an area where supervised fine-tuning falls short due to its imitation-based nature and limited exposure to diverse reasoning patterns. 
In addition to accuracy, RL achieves better performance with fewer annotated examples, since it learns from scalar rewards rather than full reasoning traces. 
This improves data efficiency and generalization to unseen multi-hop patterns.

\begin{table*}[ht]
\centering
\vspace{-0.3cm}
\caption{
Comparison of RL and SFT. (RQ1)
}
\centering
\renewcommand{\arraystretch}{1.1}
\begin{adjustbox}{width=\textwidth}
\begin{tabular}{lcccccccc}
\toprule
\multirow{2}{*}{\textbf{Method}} & \multicolumn{4}{c}{\textbf{Freebase}} & \multicolumn{4}{c}{\textbf{WikiData}} \\
\cmidrule(lr){2-5} \cmidrule(lr){6-9}
 & CWQ & WebQSP & WebQuestion & GrailQA & QALD10-en & T-REx & Zero-Shot RE & Creak \\
\midrule
Qwen-2.5-7B (SFT) + KG & 51.84 & 74.80 & 61.42 & 46.18 & 65.18 & 68.77 & 71.46 & 83.43 \\
Qwen-2.5-7B (RL) + KG & 61.07 (+9.23) & 83.20 (+8.40) & 66.90 (+5.48) & 50.10 (+3.92) & 74.28 (+9.10) & 72.14 (+3.37) & 78.64 (+7.18) & 91.82 (+8.39) \\
LLaMA-3.1-8B (SFT) + KG & 47.40 & 72.98 & 60.00 & 47.70 & 59.63 & 70.23 & 64.08 & 84.47 \\
LLaMA-3.1-8B (RL) + KG & 58.20 (+10.80) & 76.90 (+3.92) & 67.28 (+7.28) & 55.41 (+7.71) & 67.66 (+8.03) & 74.20 (+3.97) & 70.25 (+6.17) & 88.31 (+3.84) \\
\bottomrule
\end{tabular}%
\end{adjustbox}
\label{tab:sft_vs_rl}
\end{table*}

\begin{table*}[ht]
\caption{
Ablation study of the RL training process. The table reports performance across different RL variants compared to the baseline Qwen-2.5-7B (RL) + KG model. Reported (+/–) values indicate the change in Hits@1 relative to the baseline.
}
\centering
\begin{adjustbox}{width=\textwidth}
\begin{tabular}{lcccccccc}
\toprule
\multirow{2}{*}{\textbf{Method}} & \multicolumn{4}{c}{\textbf{Freebase}} & \multicolumn{4}{c}{\textbf{WikiData}} \\
\cmidrule(lr){2-5} \cmidrule(lr){6-9}
 & CWQ & WebQSP & WebQuestion & GrailQA & QALD10-en & T-REx & Zero-Shot RE & Creak \\
\midrule
{Qwen-2.5-7B (RL) + KG} & \textbf{61.07} & \underline{83.20} & \underline{66.90} & \textbf{50.10} & \textbf{74.28} & \underline{72.14} & \textbf{78.64} & \textbf{91.82} \\
\midrule
\rowcolor{gray!10}\multicolumn{9}{l}{\textbf{reward signal ablation (RQ2)}} \\
w/o Retrieval Reward    & \cellcolor{red!15}\underline{60.75} (-0.32) & \cellcolor{red!20}82.38 (-0.82) & \cellcolor{red!20}66.01 (-0.89) & \cellcolor{red!15}49.71 (-0.39) & \cellcolor{red!20}\underline{73.54} (-0.74) & \cellcolor{green!20}\textbf{72.81} (+0.67) & \cellcolor{red!20}77.81 (-0.83) & \cellcolor{red!15}91.62 (-0.20) \\
w/o Format Reward       & \cellcolor{red!20}60.45 (-0.62) & \cellcolor{green!15}\textbf{83.63} (+0.43) & \cellcolor{red!20}66.19 (-0.71) & \cellcolor{red!15}\underline{49.97} (-0.13) & \cellcolor{red!25}72.77 (-1.51) & \cellcolor{red!15}72.10 (-0.04) & \cellcolor{red!20}\underline{78.10} (-0.54) & \cellcolor{red!20}91.30 (-0.52) \\
w/o Reasoning Reward   & \cellcolor{red!35}57.94 (-3.13) & \cellcolor{red!32}80.50 (-2.70) & \cellcolor{red!32}64.42 (-2.48) & \cellcolor{red!15}49.92 (-0.18) & \cellcolor{red!32}71.69 (-2.59) & \cellcolor{red!30}70.22 (-1.92) & \cellcolor{red!35}75.75 (-2.89) & \cellcolor{red!15}\underline{91.72} (-0.10) \\
w/o Answer Reward & \cellcolor{red!50}51.96 (-9.11) & \cellcolor{red!47}75.68 (-7.52) & \cellcolor{red!50}57.15 (-9.75) & \cellcolor{red!47}42.19 (-7.91) & \cellcolor{red!50}64.81 (-9.47) & \cellcolor{red!36}68.87 (-3.27) & \cellcolor{red!45}72.31 (-6.33) & \cellcolor{red!50}82.06 (-9.76) \\
\midrule
\rowcolor{gray!10}\multicolumn{9}{l}{\textbf{without ground truth -- LLM being Judge+Evaluator (RQ3)}} \\
Evaluate Reward by GPT-4o & \cellcolor{red!47}53.49 (-7.58) & \cellcolor{red!44}77.07 (-6.13) & \cellcolor{red!50}57.46 (-9.44) & \cellcolor{red!32}47.46 (-2.64) & \cellcolor{red!38}70.53 (-3.75) & \cellcolor{red!34}69.31 (-2.83) & \cellcolor{red!42}73.82 (-4.82) & \cellcolor{red!32}89.23 (-2.59) \\
\midrule
\rowcolor{gray!10}\multicolumn{9}{l}{\textbf{without history resampling (RQ4)}} \\
w/o History Resampling  & \cellcolor{red!36}57.83 (-3.24) & \cellcolor{red!32}80.64 (-2.56) & \cellcolor{green!15}\textbf{67.19} (+0.29) & \cellcolor{red!32}47.50 (-2.60) & \cellcolor{red!30}72.16 (-2.12) & \cellcolor{red!30}70.15 (-1.99) & \cellcolor{red!25}77.39 (-1.25) & \cellcolor{red!20}91.13 (-0.69) \\
\bottomrule
\end{tabular}%
\end{adjustbox}
\label{tab:ablation}
\end{table*}

\begin{figure*}[t]
  \centering
  \vspace{-0.3cm}
  % \begin{subfigure}{0.32\textwidth}
  %   \includegraphics[width=\linewidth]{figures/reward_data.pdf}
  % \end{subfigure}\hfill
  % \begin{subfigure}{0.32\textwidth}
  %   \includegraphics[width=\linewidth]{figures/response_length.pdf}
  % \end{subfigure}\hfill
  % \begin{subfigure}{0.32\textwidth}
  %   \includegraphics[width=\linewidth]{figures/retrieval_number_0509.pdf}
  % \end{subfigure}
    \subfloat[]{
    \includegraphics[width=0.3\textwidth]{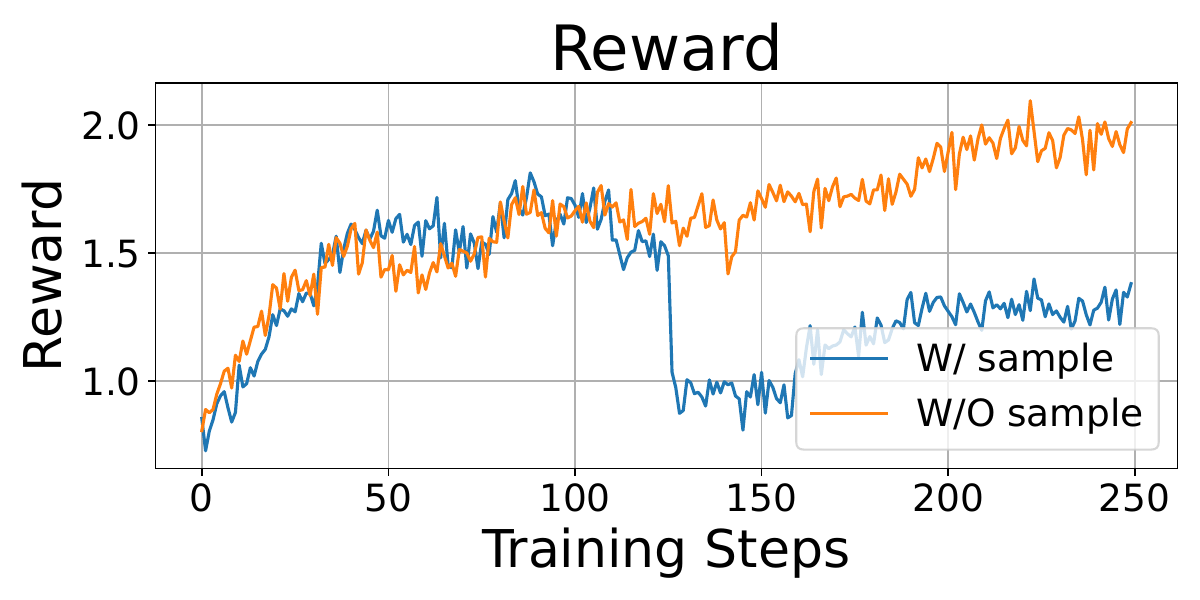}
  }\hfill
  \subfloat[]{
    \includegraphics[width=0.3\textwidth]{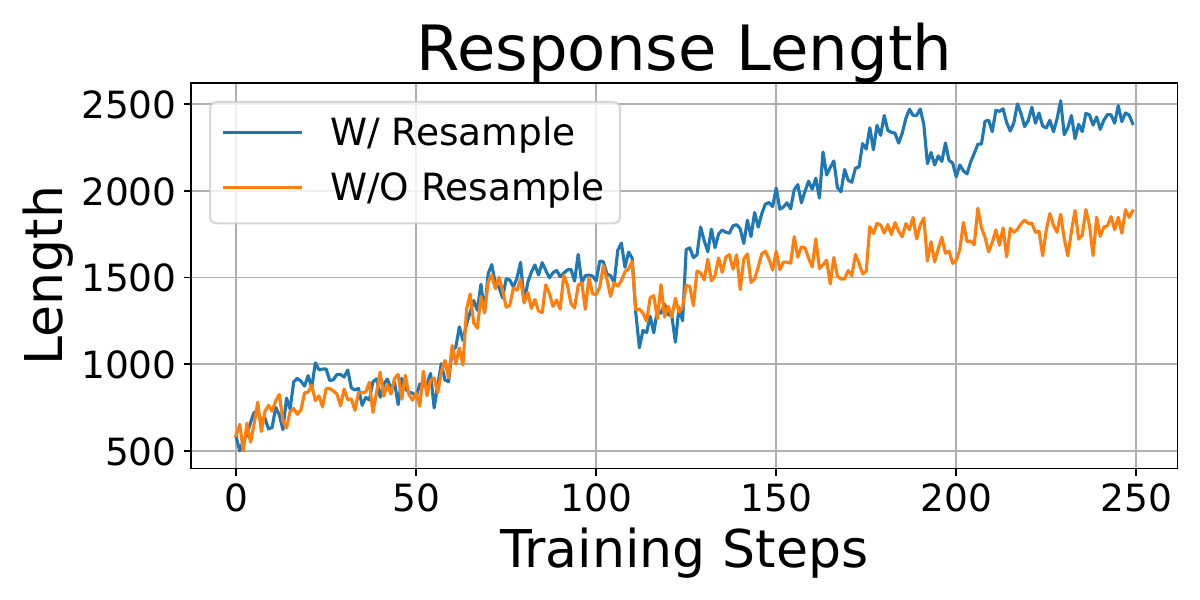}
  }\hfill
  \subfloat[]{
    \includegraphics[width=0.3\textwidth]{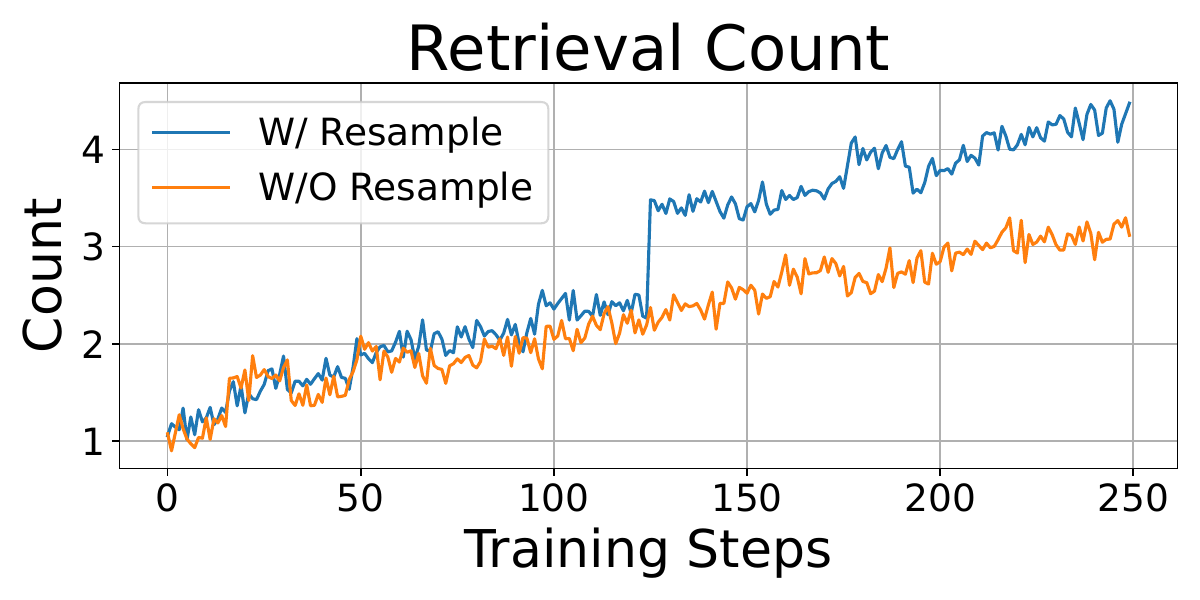}
  }
  \caption{The RL training process under two settings: with and without history resampling (\textbf{RQ4}). The figure shows how reward, response length, and retrieval count change over training steps.}
  \label{fig:three_images}
\end{figure*}

\paragraph{RL Reward Design.} We analyze the impact of individual reward components in our RL framework for KBQA. Specifically, we address the following questions:

\noindent Which components (retrieval, format, reasoning, or answer) contribute most to multi-hop KBQA performance? In particular, how do token-level action rewards (i.e. retrieval reward and format reward) compare to global sequence-level reward (i.e. reasoning and result reward)? (\textbf{RQ2})
As shown in Table~\ref{tab:ablation}, removing the reasoning reward leads to the largest performance drop across most datasets, indicating that sequence-level feedback is essential for acquiring robust, long-horizon reasoning capabilities. In contrast, the retrieval and format rewards yield smaller but consistent improvements, helping to regularize tool use and output structure. These two rewards operate at the token level, primarily guiding the generation of literal tags (e.g., \texttt{<search>}, \texttt{<answer>}) to enforce structured output and tool invocation. As such, these behaviors are effectively learned through SFT --  RL does not offer a clear efficiency advantage in this context. While token-level rewards contribute to training stability and early convergence, their gains are marginal compared to the global benefits provided by reasoning-level supervision.

\noindent For the final answer reward, can LLMs reliably serve as automated judges for final answers in RL? (\textbf{RQ3}) 
Specifically, we compare two modes: (i) \textbf{Evaluator-only}, where the LLM assesses the answer against a known ground truth, and (ii) \textbf{Judge+Evaluator}, where no ground truth is provided and the LLM both infers a reference answer and evaluates the model’s output accordingly.
Using GPT-4o in Judge+Evaluator mode results in a noticeable performance drop, though it still performs better than removing answer reward entirely. This suggests LLMs can act as fallback evaluators when ground truth is unavailable, but are less reliable than direct supervision.

% \paragraph{RL Reward Design.} We study the impact of different reward components for KBQA. 

% \noindent \textbf{RQ2:} Which rewards contribute most? 
% As shown in Table~\ref{tab:ablation}, removing the reasoning reward causes the largest performance drop, highlighting the importance of sequence-level supervision for long-horizon reasoning. In contrast, retrieval and format rewards provide smaller but consistent gains by regularizing tool usage and output structure. As token-level rewards, they mainly guide structured generation (e.g., \texttt{<search>}, \texttt{<answer>}) and can largely be learned via SFT, with limited additional benefit from RL.

% \noindent \textbf{RQ3:} Can LLMs serve as answer evaluators? 
% Using GPT-4o in Judge+Evaluator mode degrades performance compared to using ground-truth-based evaluation, though it remains better than removing answer reward. This indicates LLM-based judging is a viable fallback but less reliable than direct supervision.

\paragraph{RL Sampling Efficiency.} How effective is history resampling in mitigating overfitting to trivial or overrepresented examples during RL training? (\textbf{RQ4})
Figure~\ref{fig:three_images} presents training trajectories for three key metrics: average reward, response length, and retrieval count \cite{song2025r1}, comparing models trained with and without history resampling. All metrics show a general upward trend over time, indicating learning progress. When history resampling is introduced at the start of the second epoch, a temporary dip in average reward is observed, reflecting the removal of simpler, one-hop questions in favor of more challenging multi-hop queries. As training proceeds, the model adapts to these harder examples, resulting in a steady recovery and continued reward improvement.

Importantly, models trained with history resampling exhibit longer response lengths and higher retrieval counts, both indicative of more complex, multi-step reasoning. These results confirm that resampling effectively shifts training focus toward higher-quality samples, improving the model’s capacity for reasoning without being biased toward shallow patterns.

\section{Related Work}
\label{sec:related-work}

KBQA methods are broadly categorized as retrieval-based or semantic parsing-based. The former extract relevant KG subgraphs, while the latter translate questions into executable queries~\cite{zhang2019complex, gu-su-2022-arcaneqa}. Reinforcement learning (RL) has been used to explore multi-hop paths~\cite{xiong2017deeppath, das2018go}, though prior methods often struggle with efficiency and error accumulation. To address this, recent works incorporate LLMs for sub-question decomposition or as priors for guiding RL~\cite{zhao2024kg, zhang2025collaborative}.

% LLMs demonstrate stronger reasoning when prompted explicitly~\cite{wang2026llm,wei2022chain,wang2025automating}, and RL further enhances this via iterative reflection~\cite{zhu2025chain, guo2025deepseek}. 

Recent progress demonstrates that explicitly prompting LLMs to perform thinking before give an answer significantly enhances their problem-solving capabilities~\cite{wei2022chain, li2025system}. Reinforcement Learning has further strengthened these abilities by enabling LLMs to iteratively reflect and reason effectively~\cite{wang2024math, zhu2025chain}. While RL-augmented LLMs succeed in unstructured multi-hop retrieval~\cite{li2025webthinker}, KG reasoning remains more constrained, requiring alignment to discrete graph paths. This limits the applicability of open-ended reasoning and calls for structural-aware LLM integration.

% \section{Conclusion}
% We introduced KG-Hopper, an RL-based framework that enhances compact open LLMs for multi-hop reasoning over knowledge graphs. It integrates KG reasoning into the model’s intrinsic reasoning process, enabling multi-hop traversal and answer generation in a single inference round.
% We adopt a two-stage training paradigm with cold-start retrieval learning followed by reinforcement learning, and design reward functions, masking, and history resampling to guide efficient reasoning.
% Experiments on eight benchmarks and ablations demonstrate strong effectiveness and robustness for KBQA.

\section{Conclusion}
In this paper, we introduced KG-Hopper, a novel RL-based framework designed to empower compact open LLMs with enhanced multi-hop reasoning capabilities over KGs. To the best of our knowledge, KG-Hopper is the first framework to apply reinforcement learning to enhance multi-hop knowledge graph reasoning in large language models. Beyond this, KG-Hopper integrates KG reasoning directly into the intrinsic "thinking" process of the reasoning LLM, enabling autonomous multi-hop traversal and answer generation within a single inference round. Specifically, we propose a two-stage training paradigm: we begin with cold-start data to teach basic KG retrieval skills, and then apply reinforcement learning to substantially improve multi-hop reasoning capability. To support this, we design tailored reward functions that explicitly guide retrieval and reasoning behaviors, introduce masking techniques to control exposure to retrieved knowledge, and implement a history resampling strategy to improve training efficiency. Experimental results on eight benchmark datasets, along with comprehensive ablation studies, demonstrate the effectiveness of KG-Hopper in enabling efficient and robust KBQA.

\section*{Acknowledgment}
This work was partially funded by the Autonomous Systems and Software Program (WASP), supported by the Knut and Alice Wallenberg Foundation, and the Chalmers Artificial Intelligence Research Centre (CHAIR).

\balance
\bibliographystyle{IEEEtran}
\bibliography{references}

@article{wang2026domagent,
  title={DomAgent: Leveraging Knowledge Graphs and Case-Based Reasoning for Domain-Specific Code Generation},
  author={Wang, Shuai and Parthasarathy, Dhasarathy and Feldt, Robert and Yu, Yinan},
  journal={arXiv preprint arXiv:2603.21430},
  year={2026}
}

@article{wang2025plugging,
  title={Plugging Schema Graph into Multi-Table QA: A Human-Guided Framework for Reducing LLM Reliance},
  author={Wang, Xixi and Costa, Miguel and Kovaceva, Jordanka and Wang, Shuai and Pereira, Francisco C},
  journal={arXiv preprint arXiv:2506.04427},
  year={2025}
}

@inproceedings{wang-yu-2025-iquest,
    title = "i{QUEST}: An Iterative Question-Guided Framework for Knowledge Base Question Answering",
    author = "Wang, Shuai  and
      Yu, Yinan",
    booktitle = "Proceedings of the 63rd Annual Meeting of the Association for Computational Linguistics (Volume 1: Long Papers)",
    month = jul,
    year = "2025",
    url = "https://aclanthology.org/2025.acl-long.760/",
    doi = "10.18653/v1/2025.acl-long.760",
    pages = "15616--15628",
}

@inproceedings{onoe2creak,
  title={CREAK: A Dataset for Commonsense Reasoning over Entity Knowledge},
  author={Onoe, Yasumasa and Zhang, Michael JQ and Choi, Eunsol and Durrett, Greg},
  booktitle={Thirty-fifth Conference on Neural Information Processing Systems Datasets and Benchmarks Track (Round 2)}
}

@inproceedings{petroni2021kilt,
  title={KILT: a Benchmark for Knowledge Intensive Language Tasks},
  author={Petroni, Fabio and Piktus, Aleksandra and Fan, Angela and Lewis, Patrick and Yazdani, Majid and De Cao, Nicola and Thorne, James and Jernite, Yacine and Karpukhin, Vladimir and Maillard, Jean and others},
  booktitle={Proceedings of the 2021 Conference of the North American Chapter of the Association for Computational Linguistics: Human Language Technologies},
  pages={2523--2544},
  year={2021}
}

@inproceedings{elsahar2018t,
  title={T-rex: A large scale alignment of natural language with knowledge base triples},
  author={Elsahar, Hady and Vougiouklis, Pavlos and Remaci, Arslen and Gravier, Christophe and Hare, Jonathon and Laforest, Frederique and Simperl, Elena},
  booktitle={Proceedings of the Eleventh International Conference on Language Resources and Evaluation (LREC 2018)},
  year={2018}
}

@inproceedings{perevalov2022qald,
  title={Qald-9-plus: A multilingual dataset for question answering over dbpedia and wikidata translated by native speakers},
  author={Perevalov, Aleksandr and Diefenbach, Dennis and Usbeck, Ricardo and Both, Andreas},
  booktitle={2022 IEEE 16th International Conference on Semantic Computing (ICSC)},
  pages={229--234},
  year={2022},
  organization={IEEE}
}

@inproceedings{gu2021beyond,
  title={Beyond IID: three levels of generalization for question answering on knowledge bases},
  author={Gu, Yu and Kase, Sue and Vanni, Michelle and Sadler, Brian and Liang, Percy and Yan, Xifeng and Su, Yu},
  booktitle={Proceedings of the Web Conference 2021},
  pages={3477--3488},
  year={2021}
}

@inproceedings{berant2013semantic,
  title={Semantic parsing on freebase from question-answer pairs},
  author={Berant, Jonathan and Chou, Andrew and Frostig, Roy and Liang, Percy},
  booktitle={Proceedings of the 2013 conference on empirical methods in natural language processing},
  pages={1533--1544},
  year={2013}
}

@inproceedings{yih2016value,
  title={The value of semantic parse labeling for knowledge base question answering},
  author={Yih, Wen-tau and Richardson, Matthew and Meek, Christopher and Chang, Ming-Wei and Suh, Jina},
  booktitle={Proceedings of the 54th Annual Meeting of the Association for Computational Linguistics (Volume 2: Short Papers)},
  pages={201--206},
  year={2016}
}

@inproceedings{talmor2018web,
  title={The Web as a Knowledge-Base for Answering Complex Questions},
  author={Talmor, Alon and Berant, Jonathan},
  booktitle={Proceedings of the 2018 Conference of the North American Chapter of the Association for Computational Linguistics: Human Language Technologies, Volume 1 (Long Papers)},
  pages={641--651},
  year={2018}
}

@inproceedings{lin2018multi,
  title={Multi-Hop Knowledge Graph Reasoning with Reward Shaping},
  author={Lin, Xi Victoria and Socher, Richard and Xiong, Caiming},
  booktitle={Proceedings of the 2018 Conference on Empirical Methods in Natural Language Processing},
  pages={3243--3253},
  year={2018}
}

@inproceedings{das2018go,
  title={Go for a Walk and Arrive at the Answer: Reasoning Over Paths in Knowledge Bases using Reinforcement Learning},
  author={Das, Rajarshi and Dhuliawala, Shehzaad and Zaheer, Manzil and Vilnis, Luke and Durugkar, Ishan and Krishnamurthy, Akshay and Smola, Alex and McCallum, Andrew},
  booktitle={International Conference on Learning Representations},
  year={2018}
}

@inproceedings{xiong2017deeppath,
  title={DeepPath: A Reinforcement Learning Method for Knowledge Graph Reasoning},
  author={Xiong, Wenhan and Hoang, Thien and Wang, William Yang},
  booktitle={Proceedings of the 2017 Conference on Empirical Methods in Natural Language Processing},
  pages={564--573},
  year={2017}
}

@inproceedings{gu-su-2022-arcaneqa,
    title = "{A}rcane{QA}: Dynamic Program Induction and Contextualized Encoding for Knowledge Base Question Answering",
    author = "Gu, Yu  and
      Su, Yu",
    editor = "Calzolari, Nicoletta  and
      Huang, Chu-Ren  and
      Kim, Hansaem  and
      Pustejovsky, James  and
      Wanner, Leo  and
      Choi, Key-Sun  and
      Ryu, Pum-Mo  and
      Chen, Hsin-Hsi  and
      Donatelli, Lucia  and
      Ji, Heng  and
      Kurohashi, Sadao  and
      Paggio, Patrizia  and
      Xue, Nianwen  and
      Kim, Seokhwan  and
      Hahm, Younggyun  and
      He, Zhong  and
      Lee, Tony Kyungil  and
      Santus, Enrico  and
      Bond, Francis  and
      Na, Seung-Hoon",
    booktitle = "Proceedings of the 29th International Conference on Computational Linguistics",
    year = "2022",
    pages = "1718--1731",
}

@inproceedings{zhang2019complex,
  title={Complex question decomposition for semantic parsing},
  author={Zhang, Haoyu and Cai, Jingjing and Xu, Jianjun and Wang, Ji},
  booktitle={Proceedings of the 57th Annual Meeting of the Association for Computational Linguistics},
  pages={4477--4486},
  year={2019}
}

@article{li2025system,
  title={From system 1 to system 2: A survey of reasoning large language models},
  author={Li, Zhong-Zhi and Zhang, Duzhen and Zhang, Ming-Liang and Zhang, Jiaxin and Liu, Zengyan and Yao, Yuxuan and Xu, Haotian and Zheng, Junhao and Wang, Pei-Jie and Chen, Xiuyi and others},
  journal={arXiv preprint arXiv:2502.17419},
  year={2025}
}

@article{zhu2025chain,
  title={Chain-of-Thought Tokens are Computer Program Variables},
  author={Zhu, Fangwei and Wang, Peiyi and Sui, Zhifang},
  journal={arXiv preprint arXiv:2505.04955},
  year={2025}
}

@article{wei2022chain,
  title={Chain-of-thought prompting elicits reasoning in large language models},
  author={Wei, Jason and Wang, Xuezhi and Schuurmans, Dale and Bosma, Maarten and Xia, Fei and Chi, Ed and Le, Quoc V and Zhou, Denny and others},
  journal={Advances in neural information processing systems},
  volume={35},
  pages={24824--24837},
  year={2022}
}

@inproceedings{wang2024math,
  title={Math-Shepherd: Verify and Reinforce LLMs Step-by-step without Human Annotations},
  author={Wang, Peiyi and Li, Lei and Shao, Zhihong and Xu, Runxin and Dai, Damai and Li, Yifei and Chen, Deli and Wu, Yu and Sui, Zhifang},
  booktitle={Proceedings of the 62nd Annual Meeting of the Association for Computational Linguistics (Volume 1: Long Papers)},
  pages={9426--9439},
  year={2024}
}

@article{li2025webthinker,
  title={Webthinker: Empowering large reasoning models with deep research capability},
  author={Li, Xiaoxi and Jin, Jiajie and Dong, Guanting and Qian, Hongjin and Zhu, Yutao and Wu, Yongkang and Wen, Ji-Rong and Dou, Zhicheng},
  journal={arXiv preprint arXiv:2504.21776},
  year={2025}
}

@article{song2025r1,
  title={R1-Searcher: Incentivizing the Search Capability in LLMs via Reinforcement Learning},
  author={Song, Huatong and Jiang, Jinhao and Min, Yingqian and Chen, Jie and Chen, Zhipeng and Zhao, Wayne Xin and Fang, Lei and Wen, Ji-Rong},
  journal={arXiv preprint arXiv:2503.05592},
  year={2025}
}

@article{jaech2024openai,
  title={Openai o1 system card},
  author={Jaech, Aaron and Kalai, Adam and Lerer, Adam and Richardson, Adam and El-Kishky, Ahmed and Low, Aiden and Helyar, Alec and Madry, Aleksander and Beutel, Alex and Carney, Alex and others},
  journal={arXiv preprint arXiv:2412.16720},
  year={2024}
}

@inproceedings{zhang2025collaborative,
  title={A Collaborative Reasoning Framework Powered by Reinforcement Learning and Large Language Models for Complex Questions Answering over Knowledge Graph},
  author={Zhang, Zhiqiang and Zhao, Wen},
  booktitle={Proceedings of the 31st International Conference on Computational Linguistics},
  pages={10672--10684},
  year={2025}
}

@article{guo2025deepseek,
  title={Deepseek-r1: Incentivizing reasoning capability in llms via reinforcement learning},
  author={Guo, Daya and Yang, Dejian and Zhang, Haowei and Song, Junxiao and Zhang, Ruoyu and Xu, Runxin and Zhu, Qihao and Ma, Shirong and Wang, Peiyi and Bi, Xiao and others},
  journal={arXiv preprint arXiv:2501.12948},
  year={2025}
}

@article{zhang2025srpo,
  title={SRPO: A Cross-Domain Implementation of Large-Scale Reinforcement Learning on LLM},
  author={Zhang, Xiaojiang and Wang, Jinghui and Cheng, Zifei and Zhuang, Wenhao and Lin, Zheng and Zhang, Minglei and Wang, Shaojie and Cui, Yinghan and Wang, Chao and Peng, Junyi and others},
  journal={arXiv preprint arXiv:2504.14286},
  year={2025}
}

@article{narvekar2020curriculum,
  title={Curriculum learning for reinforcement learning domains: A framework and survey},
  author={Narvekar, Sanmit and Peng, Bei and Leonetti, Matteo and Sinapov, Jivko and Taylor, Matthew E and Stone, Peter},
  journal={Journal of Machine Learning Research},
  volume={21},
  number={181},
  pages={1--50},
  year={2020}
}

@inproceedings{
sun2024thinkongraph,
title={Think-on-Graph: Deep and Responsible Reasoning of Large Language Model on Knowledge Graph},
author={Jiashuo Sun and Chengjin Xu and Lumingyuan Tang and Saizhuo Wang and Chen Lin and Yeyun Gong and Lionel Ni and Heung-Yeung Shum and Jian Guo},
booktitle={The Twelfth International Conference on Learning Representations},
year={2024}
}

@inproceedings{xiong-etal-2024-interactive,
    title = "Interactive-{KBQA}: Multi-Turn Interactions for Knowledge Base Question Answering with Large Language Models",
    author = "Xiong, Guanming  and
      Bao, Junwei  and
      Zhao, Wen",
    editor = "Ku, Lun-Wei  and
      Martins, Andre  and
      Srikumar, Vivek",
    booktitle = "Proceedings of the 62nd Annual Meeting of the Association for Computational Linguistics (Volume 1: Long Papers)",
    month = aug,
    year = "2024",
    pages = "10561--10582",
}

@inproceedings{zhao2024kg,
  title={{KG-CoT}: Chain-of-thought prompting of large language models over knowledge graphs for knowledge-aware question answering},
  author={Zhao, Ruilin and Zhao, Feng and Wang, Long and Wang, Xianzhi and Xu, Guandong},
  booktitle={Proceedings of the Thirty-Third International Joint Conference on Artificial Intelligence (IJCAI-24)},
  pages={6642--6650},
  year={2024},
  organization={International Joint Conferences on Artificial Intelligence}
}
\end{document}